\documentclass{article}

\PassOptionsToPackage{numbers, compress}{natbib}
\usepackage[preprint]{neurips_2026}
\usepackage{pgffor}

\usepackage[utf8]{inputenc} 
\usepackage[T1]{fontenc}    
\usepackage{hyperref}       
\usepackage{url}            
\usepackage{booktabs}       
\usepackage{amsfonts}       
\usepackage{nicefrac}       
\usepackage{microtype}      
\usepackage{xcolor}         
\usepackage{graphicx}
\usepackage{multirow}
\usepackage[table]{xcolor}
\usepackage{caption}
\usepackage{hyperref}
\usepackage{amsmath}
\title{ShotCrop$^3$: Cropping Human-Centric Images into Cinematic Triple-Shot Compositions
}

\author{%
\normalfont
  Dehong Kong\textsuperscript{1,2} ,
  Lina Lei\textsuperscript{1} ,
  Lingtao Zheng\textsuperscript{1} ,
  Chenyang Wu\textsuperscript{1} ,\\
  Ailing Zhang\textsuperscript{1} ,
  Xinran Qin\textsuperscript{1} ,
  Teng Ma\textsuperscript{2} ,
  Jiaqi Xu\textsuperscript{1} ,\\
  Zhixin Wang\textsuperscript{1} ,
  Zhikai Chen\textsuperscript{1} ,
  Xuecheng Qi\textsuperscript{1} ,
  Renjing Pei\textsuperscript{1} ,
  Fan Li\textsuperscript{1}\thanks{\ \ Corresponding author}\, , \\
  \textsuperscript{1}Huawei Noah’s Ark Lab,
  \textsuperscript{2}Sun Yat-sen University, 
}

\begin{document}

\maketitle
\begin{figure}[h]
    \centering
    \includegraphics[width=0.99\textwidth]{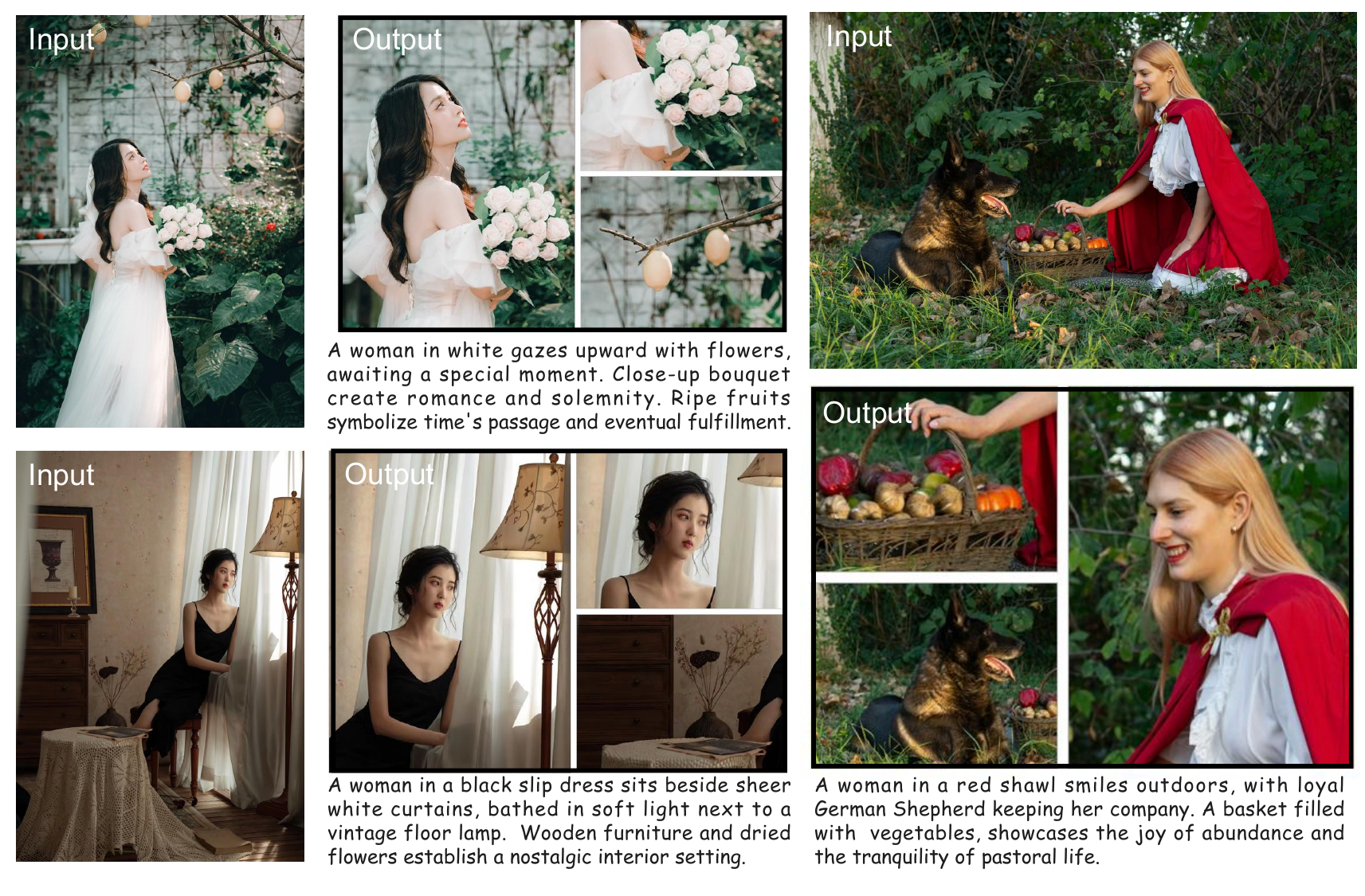}
    \caption{Given a single human-centric image, our framework outputs a cinematic triple-shot composition consisting of establishing, medium, and close-up shots with a descriptive caption}
    \label{fig:teaser}
\end{figure}

\begin{abstract}
Prior work on aesthetic composition typically produces a single aesthetically pleasing crop, overlooking the narrative value of composing multiple shots from one scene. In practice, multi-shot composition is critical for downstream creative workflows: commercial posters often require multiple crops with different emphases (e.g., context, subject, and emotion/product details) to present key story beats. Therefore, we propose \textbf{Triple-Shot Compositions (TSC)}, a composition task that generates a three-shot set~---~establishing, medium, and close-up~---~from a single human-centric image, each paired with a brief shot description to support visual narration.
To learn TSC with limited expert annotations, we introduce \textbf{ShotCrop$^3$} which  undergoes a three-stage training process: it first applies Chain-of-Thought supervised fine-tuning to establish basic reasoning and aesthetic shot-cropping skills, then performs semi-supervised fine-tuning with high-confidence pseudo labels to further enhance aesthetic  capability, and is finally optimized with Group Relative Policy Optimization for \textbf{ShotCrop$^3$} (GRPO-S) using a composite reward tailored for it. Specifically, our pseudo-labeling strategy combines MLLM-based scoring, aesthetic assessment, and CLIP similarity to retain high-confidence training signals. 
In addition, we present TSC-Bench, a benchmark of 1.2k expert-annotated test cases. Notably, ShotCrop$^3$ achieves an average improvement of \textbf{2.82} times over GPT-5 in shot localization accuracy.
\end{abstract}

\section{Introduction}
\label{sec:intro}



Recent advances~\cite{pan2024pseudo,yuan2024aesthetic,hong2024learning,liu2023beyond,nishiyasu2023image,pan2023find,shi2023joint,yang2023focusing,wang2023image} in aesthetic image cropping have primarily focused on selecting a single optimal frame from an input image.
While effective for isolated aesthetic optimization, this single-crop paradigm fails to capture the rich narrative structure inherent. In human-centric scenarios, a single crop inevitably forces a trade-off between contextual background, subject posture, and emotional details. For instance, a wide crop preserves the environment but loses facial expression, while a tight close-up captures emotion but strips away the situational context. 

In many real-world workflows—especially commercial posters and social media content—creators need multiple crops from the same scene with different narrative emphases. A single “best” crop can look appealing but often fails to jointly convey context, subject, and key details. 
By contrast, cinematic language decomposes a scene into establishing, medium, and close-up shots, each serving a distinct storytelling function. The practical value is further validated by its adoption in consumer electronics. Modern smartphone ecosystems, such as vivo’s composition tools\footnote{https://www.vivo.com.cn/vivo/x300/}, explicitly utilize a three-shot paradigm to enhance user engagement in social sharing scenarios.
Motivated by this practice, we study how to generate such multi-shot compositions from a single image.

Therefore, we propose a new aesthetic composition task, Triple-Shot Compositions (TSC), cropping human-centric images into a three-shot set—establishing, medium, and close-up—from a single image, each paired with a brief shot description to support visual narration. This multi-shot decomposition better preserves the narrative potential of the scene and produces outputs that are readily adaptable for commercial deployment and social media sharing.

Although many aesthetic cropping datasets~\cite{wei2018good,zeng2019reliable,hong2024learning,su2024spatial} have been proposed, constructing datasets for the proposed TSC task at scale remains a formidable challenge. Unlike conventional cropping, TSC annotation requires substantial domain expertise in visual storytelling to assign precise bounding boxes with cinematic intent. Annotators must possess solid composition skills and the ability to design shots. This expertise barrier makes large-scale manual supervision economically prohibitive. In addition, even state-of-the-art Multimodal Large Language Models (MLLMs)~\cite{bai2025qwen3,wang2025internvl3} struggle to perform spatially precise, narratively aware cropping without specialized, high-quality guidance.

To overcome these challenges, we first construct dataset by curating 7,600 expert-annotated image pairs, where professional photographers provide three-shot bounding boxes and leverage MLLM to generate high-fidelity captions and visual question-answering pairs for each crop. We propose \textbf{ShotCrop$^3$}, an MLLM-based framework that constructs multi-shot narratives via Triple-Shot Compositions. ShotCrop$^3$ consists of a three-stage training pipeline: (1) \textbf{Chain-of-Thought Supervised Fine-Tuning (CoT-SFT)} teaches MLLMs to establish basic reasoning about shot types (e.g., medium, close-up, and establishing shot) and aesthetic shot-cropping skills; (2) \textbf{Semi-supervised fine-tuning (Semi-SFT)} leverages the CoT-SFT model from the previous stage to generate high-confidence pseudo-labels, augmenting the dataset and further fine-tuning the model. The pseudo-label filtering strategy combines MLLM-based scoring, aesthetic assessment, and CLIP similarity to retain high-confidence training signals and ensure reliability without human intervention; (3) \textbf{Reinforcement Learning with Group Relative Policy Optimization for \textbf{ShotCrop$^3$} (GRPO-S)} applies a new reward function integrating aspect-ratio consistency, IoU-based spatial constraints, and aesthetic preferences to yield better cinematic compositions. To benchmark performance, we introduce TSC-Bench, which includes 1.2K expert-annotated images sourced from four visually diverse domains: travel photography, street photography, cinematic frames, and professional photo albums. Each image is annotated with triple bounding boxes aligned with medium, close-up, and establishing shot conventions by visual storytelling experts.

Our contributions are as follows: (1) We formalize Triple-Shot Compositions as a new task and release TSC-Bench to enable systematic evaluation; (2) We propose ShotCrop$^3$, a training framework that overcomes annotation scarcity through CoT-SFT, Semi-SFT, and GRPO-S and construct a high-quality dataset through a multi-stage process; (3) We construct a high-quality dataset and introduce a pseudo-labeling strategy that combines MLLM-based scoring, aesthetic assessment, and CLIP similarity to filter high-confidence labels; (4) Comprehensive experiments validate that ShotCrop$^3$ achieves state-of-the-art results, outperforming GPT-5.

\section{Related Work}
Recent advances in Multimodal Large Language Models (MLLMs) have demonstrated remarkable capabilities in visual understanding and reasoning. Closed-source models like Gemini 2.5 Pro\cite{comanici2025gemini} and GPT-5\cite{singh2025openai} represent the frontier of multimodal AI, offering enhanced reasoning, advanced coding skills, and multimodal understanding across text, images, and video. In the open-source domain, large-scale MLLMs such as InternVL3.5\cite{wang2025internvl3} and Qwen3-VL\cite{bai2025qwen3} have achieved competitive performance on diverse vision-language benchmarks. However, these large-scale models, despite their general capabilities, struggle with spatially precise, narratively-aware cropping tasks.


Specialist models\cite{chen2023shikra,ma2024groma,cheng2024yolo,lai2024lisa,zhong2023clipcrop,you2025photoframer,zhang2022human,ren2024dino} have been developed for specific perception, grounding, and cropping tasks.  DeepPerception\cite{ma2025deepperception} proposes a two-stage training framework combining supervised fine-tuning for cognitive reasoning scaffolding and reinforcement learning to optimize perception-cognition synergy. Recent works\cite{zhang2025procrop,qian2025zoomer,wong2025aescrop} tend to leverage MLLM for aesthetic cropping tasks.
InstructCrop\cite{sheng2025instructcrop} represents the first instruction-tuning approach for MLLM-based aesthetic image cropping. 


\section{Methodology}
\label{sec:method}

\begin{figure}[!t]
    \centering
    \includegraphics[width=0.99\textwidth]{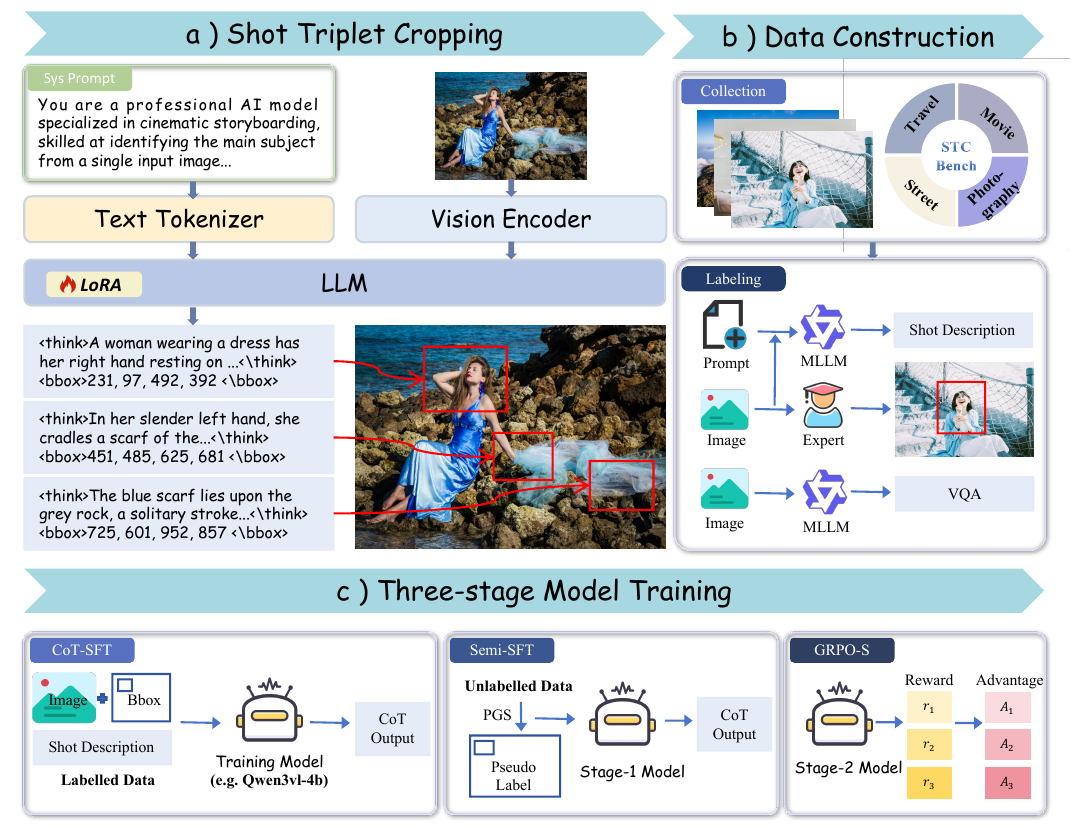}
    \caption{Overview of our method ShotCrop$^3$:a)Triple-Shot Compositions is a task that takes an image as input and outputs three shots; b)Construct TSC dataset by generate annotations from both experts and MLLMs; c)a three-stage training pipeline consists of CoT-SFT, Semi-SFT and GRPO-S.}
    \vspace{-1em}
\label{fig3}
\end{figure}

\subsection{Triple-Shot Compositions Task}
\label{subsec:TSC}

We introduce the Triple-Shot Compositions (TSC) task, cropping human-centric images into
a three-shot set—establishing, medium, and close-up—from a single image, each paired with a brief shot description to support visual narration. Specifically, given an input image $\mathcal{I}$, the model generates a three-shot set, denoted as follows:

\begin{equation}
\mathcal{O} = \{(\text{bbox}_i, \text{desc}_i)\}_{i=1}^{3},
\end{equation}

where each $\text{bbox}_i = (x_i, y_i, w_i, h_i)$ specifies the bounding box coordinates for the $i$-th crop, and $\text{desc}_i$ provides a textual description of the story content, as shown in Fig.~\ref{fig3}a).

\subsection{Data Construction}
\label{subsec:data}

To learn TSC with limited expert annotations, we construct a high-quality dataset through a multi-stage process that integrates expert knowledge and MLLM pseudo-labeling.
As depicted in Fig. \ref{fig3}b), we first curate a diverse collection of images from four distinct domains: travel photography, street photography, film stills, and professional photography portfolios.
%
%
Then, these collected images are annotated by professional photographers with over 10 years of experience in photography and cinematic production. Each image is annotated with three bounding boxes corresponding to the medium, close-up, and establishing shots, along with detailed textual descriptions for each crop that capture both aesthetic composition and narrative context. This results in a total of 7,600 image-annotation pairs, partitioned into 6,400 training samples and 1,200 test samples to ensure a rigorous evaluation protocol with no overlap between splits.
To further enrich the dataset and align with the capabilities of modern MLLMs, we leverage the Qwen3VL-235B\footnote{https://huggingface.co/Qwen/Qwen3-VL-235B-A22B-Instruct} model to generate high-fidelity captions for the TSC-Bench benchmark. The captions are designed to capture both the visual content and the narrative context of each crop, enabling the model to learn the semantic relationships between the original image and its cropped variants. More details are in Appendix Fig. \ref{figdata}.

\subsection{Three-stage Model Training}
\label{subsec:training}

As depicted in Fig. \ref{fig3}c), our training methodology comprises three sequential stages, each designed to progressively refine the model's ability to generate high-quality Triple-Shots through a combination of supervised learning, semi-supervised learning, and reinforcement learning.

\textbf{Stage 1: Chain-of-Thought Supervised Fine-Tuning (CoT-SFT).}
In the first stage, we adopt a pre-trained MLLM and apply Chain-of-Thought supervised fine-tuning on our constructed dataset $\mathcal{D}_{\text{CoT-SFT}}$, as detailed in Section~\ref{subsec:data}. The training objective is to maximize the likelihood of generating the correct CoT reasoning and corresponding bounding boxes given the input image and system prompt. This stage establishes a strong foundation for the model to learn the mapping between visual inputs and structured CoT outputs, enabling it to generate accurate bounding boxes and detailed reasoning. We also supervised finetune the MLLM without Chain-of-Thought on $\mathcal{D}_{\text{SFT}}$ for Pseudo-label Generation.


\textbf{Stage 2: Semi-Supervised Fine-Tuning (Semi-SFT).}
Building upon the Stage 1 model, we generate candidate crops for a large pool of unlabeled images. To ensure the quality of these candidates, we employ a pseudo-label generation strategy (detailed in Section~\ref{subsec:pseudo}), which filters high-confidence proposals and identifies hard examples that require manual annotation. The pseudo-labeled data are then combined with the Stage 1 training data to form the Stage 2 augmented dataset:

\begin{equation}
\mathcal{D}_{\text{Semi-SFT}} = \mathcal{D}_{\text{CoT-SFT}} \cup \mathcal{D}_{\text{pseudo}}.
\end{equation}

We further fine-tune the CoT-SFT model on $\mathcal{D}_{\text{Semi-SFT}}$ using the same objective as Stage 1 CoT-SFT, enhancing the reasoning and aesthetic shot-cropping skills. This semi-supervised approach significantly expands the effective training data while maintaining quality control, enabling the model to generalize to a wider range of visual compositions and narrative contexts.

\textbf{Stage 3: Reinforcement Learning via GRPO.}
Following the two-stage fine-tuning, we conduct GRPO-S using a separate subset of training data to further enhance the model's perception capabilities. GRPO-S trains the agent with three task-specific rewards. For a generated trajectory $\tau$ containing predicted bounding boxes ${\hat{\mathbf{b}}i}{i=1}^{3}$, where $\hat{\mathbf{b}}i = (x_i, y_i, w_i, h_i)$, the composite reward is formulated as

\begin{equation}
    R(\tau) = \lambda_{\text{IoU}} \, R_{\text{IoU}}(\tau)
+ \lambda_{\text{aes}} \, R_{\text{aesthetic}}(\tau)
+ \lambda_{\text{ratio}} \, R_{\text{ratio}}(\tau),
\end{equation}

with weights $\lambda_{\text{IoU}} = 0.6$, $\lambda_{\text{aes}} = 0.2$, and $\lambda_{\text{ratio}} = 0.2$ determined through validation set ablation.

\textbf{IoU reward.}
The IoU reward measures geometric alignment between the predicted bounding box $\hat{\mathbf{b}}$ and the ground-truth box $\mathbf{b}^{\text{gt}}$. The reward is computed as
\begin{equation}
    R_{\text{IoU}}(\hat{\mathbf{b}}, \mathbf{b}^{\text{gt}})
= \frac{|\hat{\mathbf{b}} \cap \mathbf{b}^{\text{gt}}|}
{|\hat{\mathbf{b}} \cup \mathbf{b}^{\text{gt}}|},
\end{equation}
where $|\cdot|$ denotes area.

\textbf{Aesthetic reward.}
The aesthetic reward quantifies visual quality using the Aesthetic Scorer. For a cropped region $\mathcal{I}^{\text{crop}} = \mathrm{Crop}(\mathcal{I}, \hat{\mathbf{b}})$, we obtain an aesthetic score $a \in [0,1]$:
\begin{equation}
R_{\text{aesthetic}}(\mathcal{I}^{\text{crop}}) = a.
\end{equation}

\textbf{Aspect ratio reward.}
Relying solely on the IoU and aesthetic rewards may result in extreme aspect ratios. The aspect ratio reward enforces cinematic framing conventions by penalizing deviations from standard aspect ratios (4:3 for horizontal compositions and 3:4 for vertical compositions). For a predicted box with aspect ratio $r = w/h$, we define
\begin{equation}
    R_{\text{ratio}}(r) =
\begin{cases}
\max\!\left(0,\; 1 - \left|\log r - \log \tfrac{4}{3}\right|\right), & r \ge 1,\\[4pt]
\max\!\left(0,\; 1 - \left|\log r - \log \tfrac{3}{4}\right|\right), & r < 1.
\end{cases}
\end{equation}

This reward equals 1 when the predicted ratio matches one of the target ratios, and decreases as the log-space deviation grows. The $\max(0,\cdot)$ truncation prevents negative values, keeping the reward bounded in $[0,1]$, and effectively assigns zero reward when the aspect ratio deviates too far from all targets.

\subsection{Pseudo-label Generation Strategy}
\label{subsec:pseudo}
\begin{figure}[!t]
    \centering
    \includegraphics[width=0.99\textwidth]{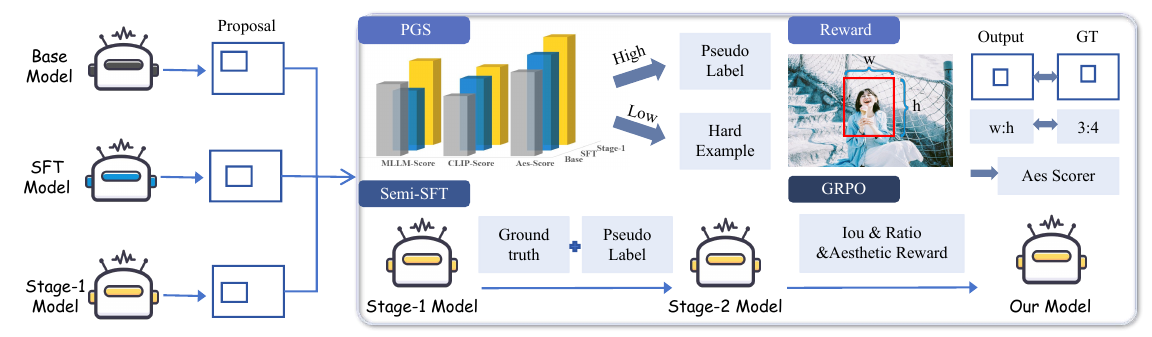}
    \caption{Details of pseudo-label generation strategy and reward function.}
    \vspace{-1em}
\label{fig4}
\end{figure}
To effectively leverage unlabeled data in Stage 2, we design a pseudo-label generation strategy (Fig. \ref{fig4}) that combines multiple evaluation metrics to ensure high-quality annotations. Given an unlabeled image $\mathcal{I}{\text{unl}}$, we first generate three candidate crops (proposals) using three distinct models: the pre-trained MLLM, the CoT-SFT model from Stage 1, and an intermediate SFT model trained on $\mathcal{D}_{\text{SFT}}$. Each model produces a set of candidate bounding boxes:

\begin{equation}
\mathbf{B}_i = \{\text{bbox}_{i1}, \text{bbox}_{i2}, \text{bbox}_{i3}\}, \quad i \in \{\text{base}, \text{SFT}, \text{CoT-SFT}\}.
\end{equation}

We then evaluate each proposal using three scoring mechanisms:
\begin{itemize}
\item \textbf{MLLM-based scoring $S_{\text{MLLM}} \in [0,1]$:} Assesses the semantic consistency between the crop and the original image, based on the relation of the crop's content to the overall narrative via an MLLM (different from the SFT model).
\item \textbf{CLIP-based scoring $S_{\text{CLIP}} \in [0,1]$:} Computes the CLIP score between the cropped image and the text description of the shot type (e.g., "medium shot") to measure the accuracy of the shot composition.
\item \textbf{Aesthetic scoring $S_{\text{aesthetic}} \in [0,1]$:} Uses a pre-trained aesthetic assessment model~\footnote{https://huggingface.co/rsinema/aesthetic-scorer} to evaluate the visual quality, such as composition, lighting, and color harmony.
\end{itemize}

For each shot, we evaluate the three candidate crops. The one selected as a pseudo-label must satisfy two criteria: (1) all three scores are higher than those of the other two proposals, and (2) each score exceeds a predefined threshold $\tau_{\text{high}}$. This ensures that only high-confidence, high-quality proposals are incorporated into the augmented training data.

%

Conversely, if all three proposals for a shot type have scores below a lower threshold $\tau_{\text{low}}$, the image is flagged as a \textit{hard example}. These hard examples are prioritized for manual annotation by our expert photographers, ensuring that challenging cases are accurately labeled and included in the augmented training set. This strategy balances data efficiency with annotation quality, enabling the model to learn from both easy and hard examples. Additionally, we employ a confidence calibration mechanism that dynamically adjusts $\tau_{\text{high}}$ and $\tau_{\text{low}}$ based on the model's performance on the validation set, ensuring that the pseudo-labeling process remains robust throughout training.

\section{Experiment}
\subsection{Experimental Setup}
\paragraph{Datasets.}
To facilitate the Triple-Shot Compositions (TSC) task, we introduce TSC-Bench  designed to evaluate storytelling oriented cropping task. As detailed in Section 3.2, we propose dataset comprises 7,600 pairs sourced from four diverse domains: travel photography, street photography, cinematic frames, and professional photography. This diversity ensures robustness against compositional styles. The dataset is partitioned into 6,400 training samples and 1,200 test samples. 

\paragraph{Baseline Models.}
Closed-Source General MLLMs: We include state-of-the-art proprietary models, Gemini 2.5 Pro\cite{comanici2025gemini} and GPT-5\cite{singh2025openai}, using zero-shot prompting with detailed cinematic instructions. These represent the upper bound of general visual reasoning capabilities.
Large-Scale Open-Source MLLMs: We evaluate InternVL3.5-38B\cite{wang2025internvl3} and Qwen3-VL-32B\cite{bai2025qwen3}.
Specialist Cropping Models: We compare with specialist cropping methods, including DeepPerception\cite{ma2025deepperception}, and InstructCrop\cite{sheng2025instructcrop}]. 
4B-Scale MLLMs: We include InternVL3.5-4B\cite{wang2025internvl3} and the base Qwen3-VL-4B\cite{bai2025qwen3}.

\paragraph{Evaluation Settings.}
Intersection-over-Union (IoU): We calculate the IoU between predicted and ground-truth bounding boxes.
Boundary Displacement Error (BDE): We measure the average Euclidean distance between the boundaries of the predicted and ground-truth boxes.
Unipercent Score: We utilize the Unipercent\cite{cao2025unipercept} to score the Image Aesthetics Assessment (IAA), Image Quality Assessment (IQA), and Image Structure \& Texture Assessment (ISTA).
Overall score:  Gemini-based scores to evaluate aesthetic
quality and storytelling ability.

\subsection{Main Results}

\begin{table*}[t]
\caption{TSC results of our model and baseline models. All metrics are averaged over Middle, Close-up, and Establishing shots. }
\tiny
\centering
\resizebox{\textwidth}{!}{
\begin{tabular}{l|c|c|ccc|cc}
\toprule
\multirow{2}{*}{\textbf{Models}} & 
\multirow{2}{*}{\textbf{IoU$\uparrow$}} & 
\multirow{2}{*}{\textbf{BDE$\downarrow$}} & 
\multicolumn{3}{c|}{\textbf{Unipercent$\uparrow$}} & 
\multicolumn{2}{c}{\textbf{Overall$\uparrow$}} \\
\cmidrule(lr){4-6}\cmidrule(lr){7-8}
& & & \textbf{IAA} & \textbf{IQA} & \textbf{ISTA} & \textbf{aesthetic} & \textbf{storytelling} \\
\midrule
\multicolumn{8}{c}{\textbf{Closed-source Models}} \\
\midrule
Gemini2.5 pro\cite{comanici2025gemini} & 0.187 & 0.246 & 0.547 & 0.575 & 0.437 & 0.815 & 0.621 \\
gpt-5\cite{singh2025openai} & 0.168 & 0.232 & 0.550 & 0.586 & 0.438 & 0.810 & 0.609 \\
\midrule
\multicolumn{8}{c}{\textbf{Large-Scale MLLMs}} \\
\midrule
InternVL3.5-38B\cite{wang2025internvl3} & 0.238 & 0.170 & 0.532 & 0.562 & 0.415 & 0.808 & 0.617 \\
Qwen3-VL-32B\cite{bai2025qwen3} & 0.312 & 0.150 & 0.511 & 0.560 & 0.415 & 0.809 & 0.580 \\
\midrule
\multicolumn{8}{c}{\textbf{Specialist Models}} \\
\midrule
DeepPerception\cite{ma2025deepperception} & 0.268 & 0.168 & 0.499 & 0.547 & 0.405 & 0.788 & 0.511 \\
InstructCrop\cite{sheng2025instructcrop} & 0.252 & 0.180 & 0.544 & 0.589 & 0.451 & 0.824 & 0.371 \\
\midrule
\multicolumn{8}{c}{\textbf{4b-Scale MLLMs}} \\
\midrule
InternVL3.5-4B\cite{wang2025internvl3} & 0.141 & 0.204 & 0.490 & 0.503 & 0.367 & 0.723 & 0.465 \\
InternVL3.5-4B (sft) & 0.424 & 0.120 & 0.547 & 0.593 & 0.448 & 0.813 & 0.582 \\
Qwen3-VL-4\cite{bai2025qwen3} & 0.292 & 0.158 & 0.505 & 0.551 & 0.412 & 0.793 & 0.521 \\ 
Qwen3-VL-4B (sft) & 0.457 & 0.106 & 0.543 & 0.593 & 0.449 & 0.812 & 0.578 \\
\midrule
\rowcolor{blue!10}
Ours & \textbf{0.544} & \textbf{0.087} & \textbf{0.554} & \textbf{0.600} & \textbf{0.455} & \textbf{0.826} & \textbf{0.623} \\
\bottomrule
\end{tabular}
}
\vspace{-1em}
\label{tab:main_result}
\end{table*}

Table \ref{tab:main_result} presents the quantitative comparison on TSC-Bench. Our ShotCrop$^3$ achieves state-of-the-art performance across all metrics, demonstrating the effectiveness of our three-stage training framework. 
\begin{figure}[!t]
    \centering
    \includegraphics[width=0.99\textwidth]{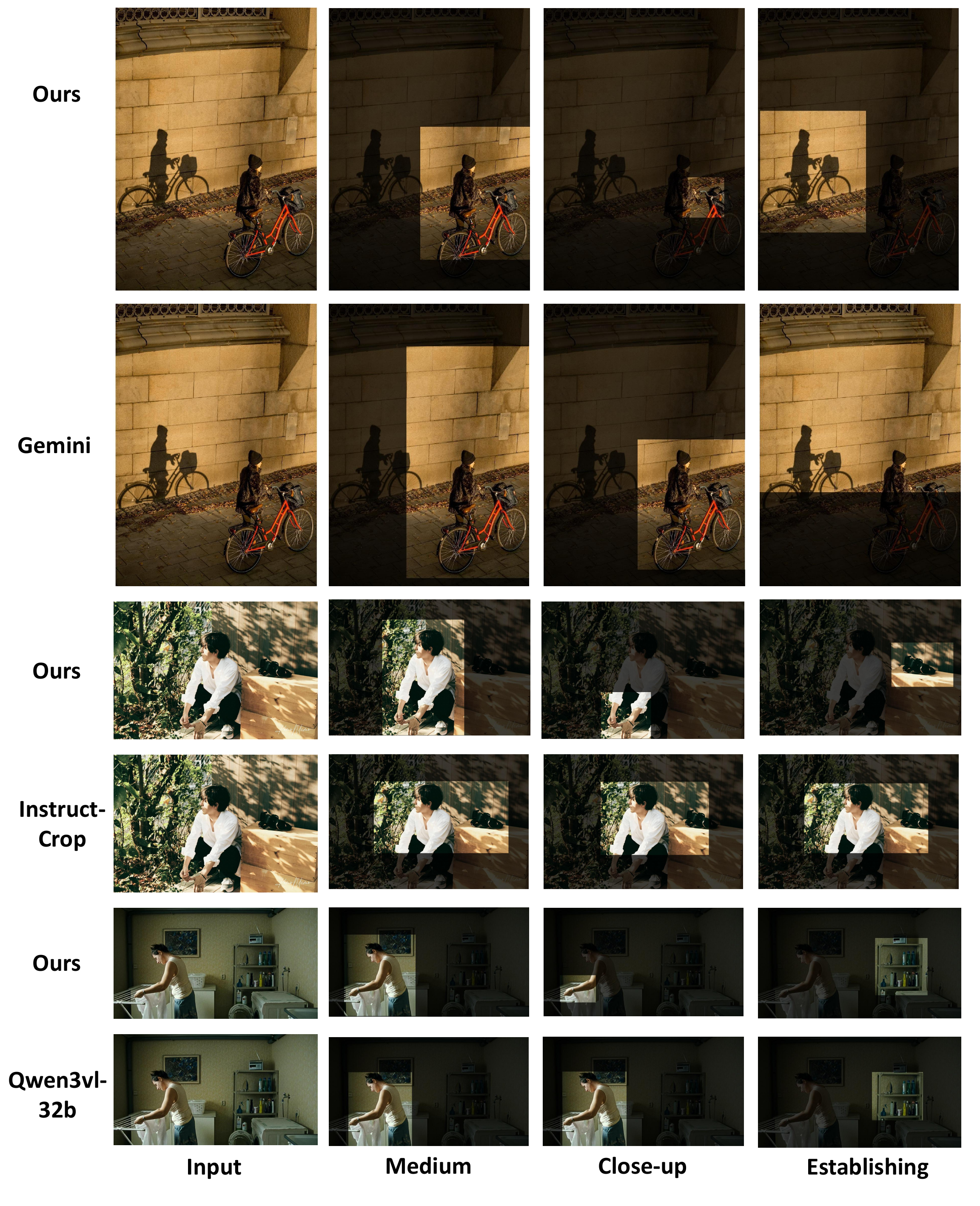}
     \vspace{-1em}
    \caption{Qualitative comparison on TSC with baseline models.}
    \vspace{-1em}
\label{vis}
\end{figure}
While traditional metrics like IoU and BDE focus on geometric alignment, and Unipercent metrics assess visual quality, they lack the capacity to evaluate TSC for visual storytelling. the Storytelling score specifically quantifies the semantic consistency, causal logic, and emotional flow across the generated triple-shots. Results shows the triple-shots cropped by other methods lack effective storytelling capability.

 Despite having only 4B parameters, ShotCrop$^3$ outperforms Qwen3-VL-32B (8$\times$ larger) by \textbf{74.3\%} in average IoU and \textbf{7.4\%} in Overall Score. Even more remarkably, ShotCrop$^3$ surpasses the fine-tuned version of Qwen3-VL-32B, demonstrating that our task-specific training strategy can compensate for and even exceed the advantages of parameter scale. Specialist models like InstructCrop and deepPerception leverages MLLM to generate single aesthetic image and grounding. However, they perform poorly on the TSC task. 
 

\subsection{Qualitative Analysis}
As illustrated in Fig. \ref{vis}, we provide a qualitative comparison between our method and state-of-the-art models. Compared to other methods, our approach demonstrates significantly higher aesthetic quality across three shots. Some methods like Gemini, which occasionally produce unbalanced frame and incomplete subject. Benefiting from the aesthetic and ratio rewards, our method fundamentally adheres to principles of aesthetic composition.
In terms of visual storytelling, our method demonstrates a superior understanding of cinematic TSC, particularly in its strict adherence to specific shot types and narrative construction. 


\subsection{Discussion}
\begin{figure}[!t]
    \centering
    \includegraphics[width=0.99\textwidth]{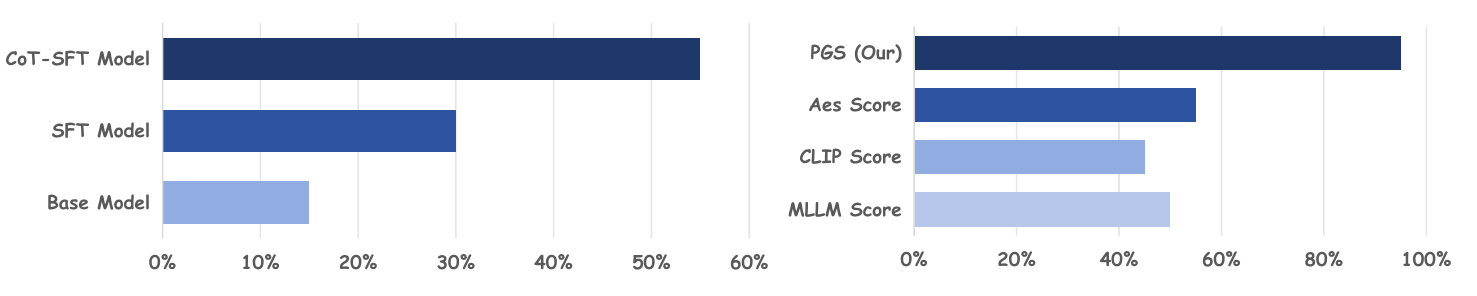}
    \caption{In the pseudo-label generation strategy, the selection rates of different models (left) and the performance of different evaluation models (right).}
    \vspace{-1em}
\label{figdis}
\end{figure}

In this section, we analyzed the contribution ratio of final selected pseudo-labels from each model variant. As shown in Figure~\ref{figdis} left, the Base Model, SFT Model, and CoT-SFT Model all contributed accepted pseudo-labels. This distribution indicates that although the CoT-SFT model has strongest ability, It fails to achieve the best performance across all cases. By aggregating proposals from all three models, we leverage their complementary strengths rather than relying on a single source.
We further investigated the impact of different scoring combinations on pseudo-label quality. When employing a single scoring standard (e.g., MLLM semantic score or CLIP similarity alone), the selection accuracy remained suboptimal due to the unidimensional nature of the evaluation, which often fails to filter out noisy samples that score highly on one metric but lack overall quality. 
\subsection{Ablation Study}


To validate the contribution of each stages in our framework, we conduct ablation studies (table \ref{tab:ablation}) on TSC-Bench. Adding CoT-SFT (Stage 1) results in a 15.88\% improvement in MLLM Score, confirming that our constructed dataset achieves favorable performance in Supervised Fine-Tuning. Introducing the pseudo-label generation strategy (Stage 2) increases IoU by 2.76\%, demonstrating the effectiveness of our method in expanding high-quality training data. Most critically, adding the GRPO stage (Stage 3) leads to a 2.98\% improvement in Story Score, validating that reward-aware reinforcement learning is essential for enhancing storytelling. 


\begin{table}[t]
\tiny
    \centering
    \caption{Ablation study on TSC-Bench about training stages and rewards. Checkmarks ($\checkmark$) indicate the inclusion of specific components.}
    \label{tab:ablation}
    \resizebox{\columnwidth}{!}{%
    \begin{tabular}{l ccccc cccc}
        \toprule
        \multirow{2}{*}{\textbf{Model}} & \multicolumn{5}{c|}{\textbf{Components}} & \multicolumn{4}{c}{\textbf{Metrics}} \\
        \cmidrule(lr){2-6} \cmidrule(l){7-10}
         & S-1 & S-2 & $R_{IoU}$ & $R_{rat}$ & $R_{aes}$ & \textbf{IoU$\uparrow$} & \textbf{BDE$\downarrow$} & \textbf{Unipercent$\uparrow$} & \textbf{Overall$\uparrow$} \\
        \midrule
        \rowcolor{gray!10}
        Base (SFT) & & & & & & 0.457 & 0.106 & 0.528 & 0.578 \\
        Base & & & & & & 0.293 & 0.159 & 0.490 & 0.422 \\
        + CoT-SFT & $\checkmark$ & & & & & 0.498 & 0.096 & 0.529 & 0.695 \\
        + Semi-SFT & $\checkmark$ & $\checkmark$ & & & & 0.512 & 0.093 & 0.531 & 0.702 \\
        \midrule
        + $R_{IoU}$ & $\checkmark$ & $\checkmark$ & $\checkmark$ & & & 0.545 & 0.088 & 0.534 & 0.718 \\
        + $R_{ratio}$ & $\checkmark$ & $\checkmark$ & $\checkmark$ & $\checkmark$ & & 0.545 & 0.088 & 0.535 & 0.721 \\
        \rowcolor{blue!10}
        + $R_{aes}$ & $\checkmark$ & $\checkmark$ & $\checkmark$ & $\checkmark$ & $\checkmark$ & \textbf{0.545} & \textbf{0.088} & \textbf{0.537} & \textbf{0.725} \\
        \bottomrule
    \end{tabular}%
    }
\end{table}
\begin{figure}[!t]
    \centering
    \includegraphics[width=0.99\textwidth]{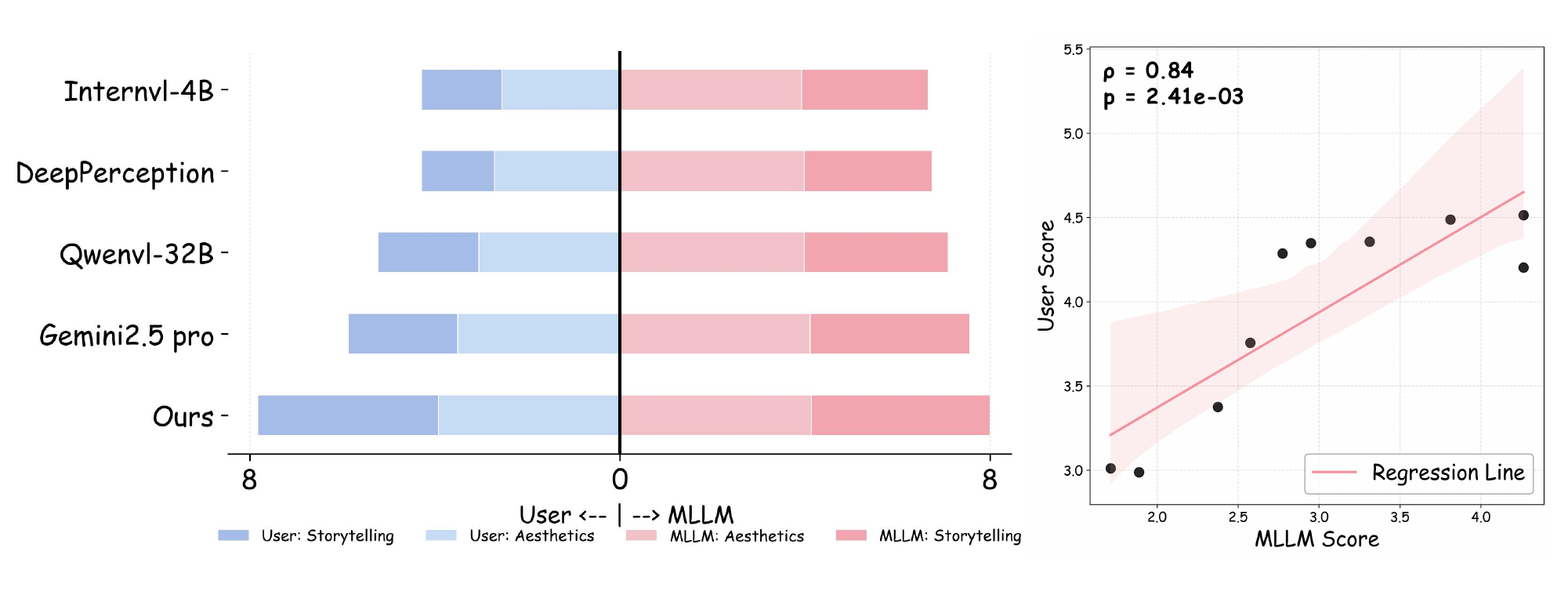}
    \caption{Comparison between user score and MLLM score (left) and analyzation on spearman rank correlation (right).}
\label{user}
\end{figure}
\subsection{User Study}
To evaluate the effectiveness of our framework, we conducted the user study involving three groups of participants: professional photographers, technical experts, and general users. The study aimed to assess the subjective user experience in terms of aesthetics and storytelling, compared with our MLLM scores. As shown in Fig. \ref{user}, the left plot presents a comparison of user ratings across different models, with separate bars for user aesthetics (blue) and user storytelling (red). It demonstrate our superiority in both user and MLLM. The right plot visualizes the strong correlation between the user scores and Overall scores, analyzed through Spearman's rank correlation.
The correlation coefficient of 0.84, with a p-value of 2.41e-03, indicates a statistically significant positive relationship, confirming that the subjective evaluations from users align well with the quantitative MLLM results.

\section{Conclusion}
In this paper, we formalized Triple-Shot Compositions (TSC), a novel task to crop human-centric Images into three cinematic crops. By introducing the ShotCrop$^3$ framework, we addressed the challenges of expert-level annotation through a three-stage pipeline. Our results on the proposed TSC-Bench demonstrate that ShotCrop$^3$ outperforms state-of-the-art MLLMs like Gemini. This work is more practical than a single aesthetic composition for commercial poster production and social media sharing. while our current framework focuses on static human-centric images, extending $ShotCrop^3$ to video domains represents a promising direction for future work.

\clearpage
{

\small
\bibliography{main}
\bibliographystyle{plain}
}
\newpage

\appendix

\renewcommand{\thesection}{\Alph{section}}
\section{Qualitative Comparison}
Figures 9-16 present additional qualitative comparisons from TSC-Bench, highlighting the superiority of our ShotCrop$^3$ in terms of aesthetic quality and visual narratives. 
\section{System Prompt of Our Data}
The prompt template is shown in Figure 17. This prompt defines three different shots in detail and imposes requirements on the MLLM regarding aesthetics, aspect ratio, and more.
\section{Prompt for MLLM-base Metric}
Prompt for MLLM-base Metric is shown in Figure 18. This prompt is designed to evaluate the ability to assess visual narratives through three shots. It features a two-dimensional scoring system—aesthetic quality and storytelling —each with detailed criteria and a 10-point scale.

\section{Details of Answer with CoT}
8 examples of answer with CoT is shown in Figures 19-22. Each example consists of three alternating <think> and <bbox> tags. The <think> tags contain natural language descriptions of visual elements—such as a woman holding tulips by bamboo blinds or a figure by a window with a coffee cup—while the <bbox> tags provide corresponding spatial coordinates for the mentioned subjects.

\section{Implementation Details}
We adopt 8 NVIDIA H100 GPUs with 80G of memory. 
We use lora in stage-1 and stage-2 with lora rank as 8 and lora alpha as 32. We set learning rate as $1e-5$ and accumulated batch size as 16.
The GRPO stage employs a maximum completion length of 8192 tokens, beta of 0.001 and number generations of 8 per input. In Pseudo-label Generation Strategy, $\tau_{\text{low}}$ and  $\tau_{\text{high}}$ is set to 0.6 and 0.85.

\foreach \n in {1,2,3,4,5,6,7,8} {  
    \begin{figure}[h]
        \centering
        \includegraphics[page=\n, scale=0.65]{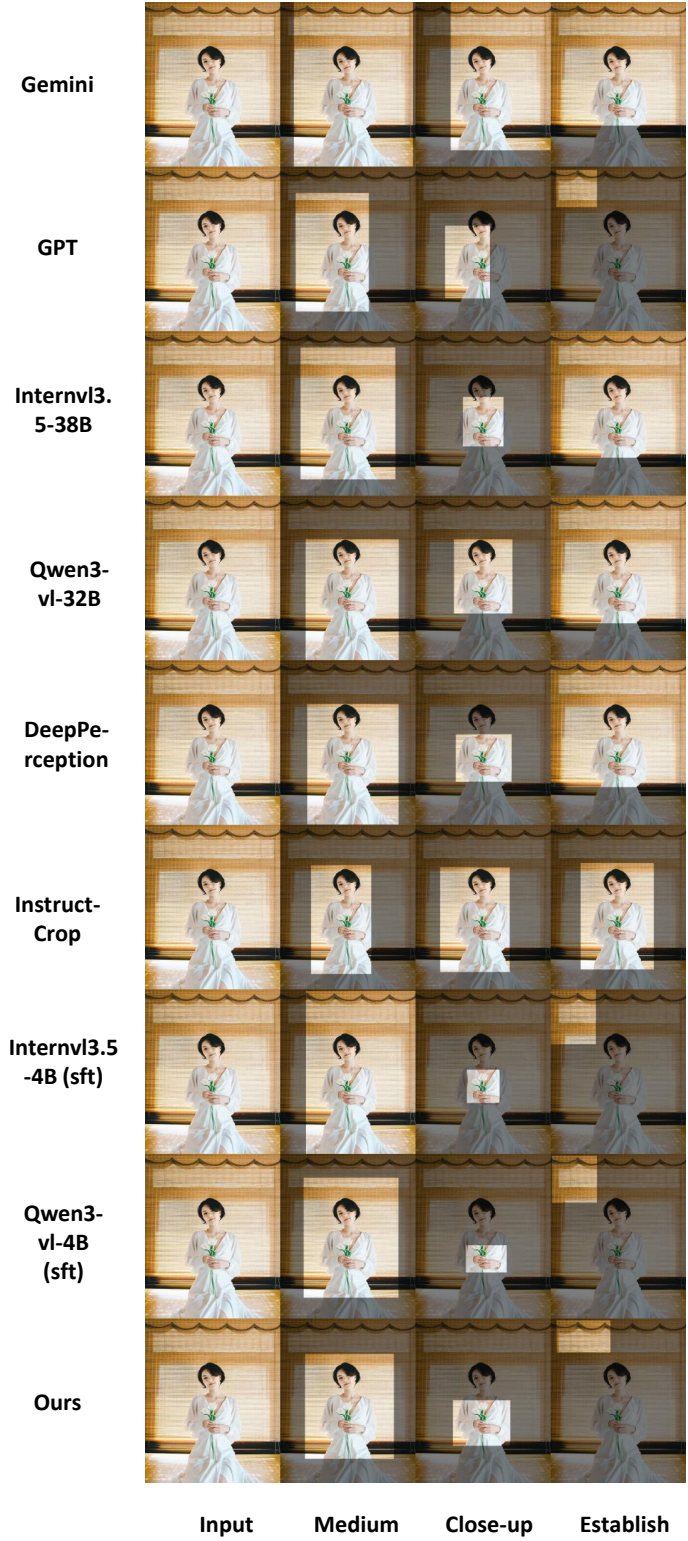}
        \caption{Example \n\ of qualitative comparison on TSC with baseline models.}
        \label{fig:pdfpage\n}
    \end{figure}
    \clearpage
}

\foreach \n in {1,2,3,4,5,6} {
    \begin{figure}[h]
        \centering
        \includegraphics[page=\n, scale=0.85]{fig/supple_cot.pdf}
        \caption{%
            \ifnum\n=1 System prompt of our STC-Bench.\fi
            \ifnum\n=2 MLLM scorer prompt for evaluating aesthetic quality and storytelling ability.\fi
            \ifnum\n=3 Answer with CoT of example 1 and example 2. \fi
            \ifnum\n=4 Answer with CoT of example 3 and example 4.\fi
            \ifnum\n=5 Answer with CoT of example 5 and example 6.\fi
            \ifnum\n=6 Answer with CoT of example 6 and example 8.\fi
        }
        \label{fig:pdfpage\n}
    \end{figure}
    \clearpage
}

\begin{figure}[!t]
    \centering
    \includegraphics[width=0.99\textwidth]{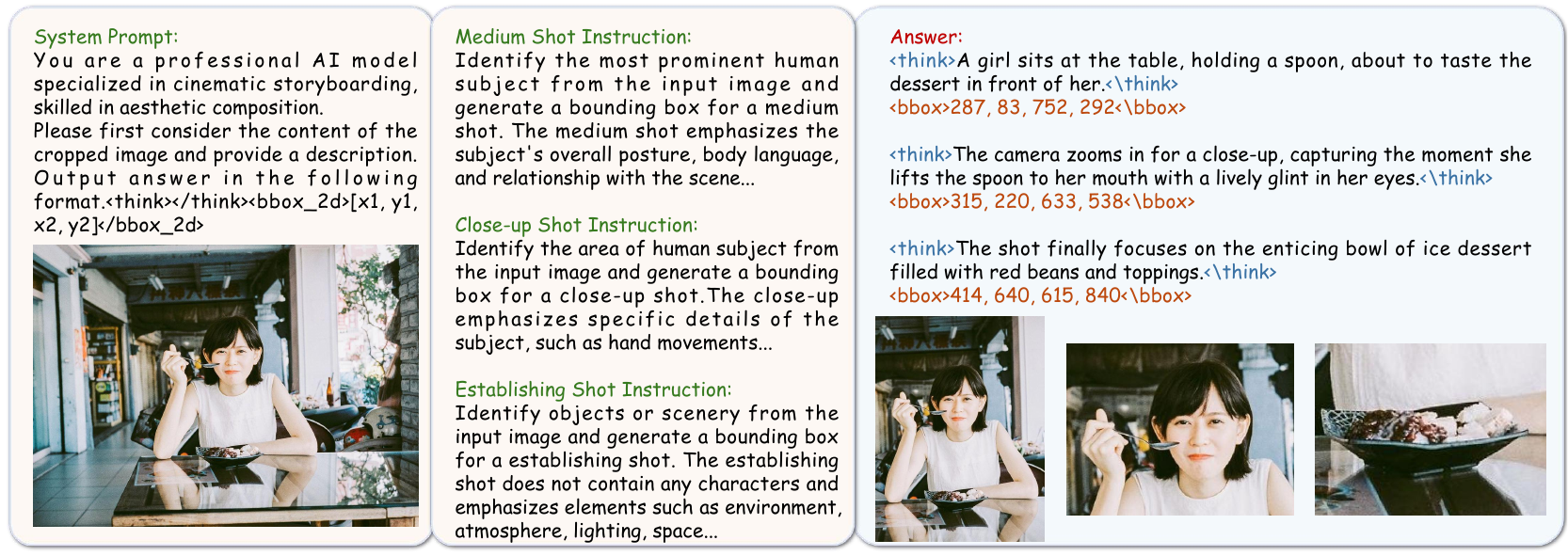}
    \caption{An example of our datasets.}
    \vspace{-1em}
\label{figdata}
\end{figure}

 \clearpage


\end{document}